\documentclass[11pt,a4paper]{article}
\usepackage[hyperref]{acl2020}
\usepackage{times}
\usepackage{latexsym}

\usepackage{microtype}

\usepackage{amsmath}
\usepackage{amssymb}
\usepackage{color,colortbl}
\usepackage{booktabs}
\usepackage[shortlabels]{enumitem}
\usepackage{subcaption}
\usepackage{multirow}
\usepackage{ifthen}
\usepackage{xspace}
\usepackage{graphicx}
\usepackage{framed}
\usepackage{textcomp}
\usepackage{float}
\usepackage{makecell}
\usepackage{booktabs}

\usepackage{url}

\aclfinalcopy 

\newif\ifcomment
\commenttrue

\newcommand{\1}{\mathbb{I}} 

\newcommand\refsec[1]{Section~\ref{sec:#1}}

\newcommand\reffig[1]{Figure~\ref{fig:#1}}

\newcommand\reftab[1]{Table~\ref{tab:#1}}
\newcommand\refapp[1]{Appendix~\ref{sec:#1}}

\ifthenelse{\isundefined{\definition}}{\newtheorem{definition}{Definition}}{}
\ifthenelse{\isundefined{\assumption}}{\newtheorem{assumption}{Assumption}}{}
\ifthenelse{\isundefined{\hypothesis}}{\newtheorem{hypothesis}{Hypothesis}}{}
\ifthenelse{\isundefined{\proposition}}{\newtheorem{proposition}{Proposition}}{}
\ifthenelse{\isundefined{\theorem}}{\newtheorem{theorem}{Theorem}}{}
\ifthenelse{\isundefined{\lemma}}{\newtheorem{lemma}{Lemma}}{}
\ifthenelse{\isundefined{\corollary}}{\newtheorem{corollary}{Corollary}}{}
\ifthenelse{\isundefined{\alg}}{\newtheorem{alg}{Algorithm}}{}
\ifthenelse{\isundefined{\example}}{\newtheorem{example}{Example}}{}

\definecolor{darkgreen}{rgb}{0,0.5,0}
\ifcomment
\newcommand\pl[1]{\textcolor{red}{[PL: #1]}}

\newcommand\rj[1]{\textcolor{blue}{[RJ: #1]}}
\newcommand\ak[1]{\textcolor{darkgreen}{[AK: #1]}}
\else
\newcommand\pl[1]{}
\newcommand\rj[1]{}
\newcommand\ak[1]{}
\fi

\newcommand\nl[1]{``\textit{#1}''}
\newcommand\psource{p_\text{source}}
\newcommand\pknownoutlier{q_\text{known}}
\newcommand\punknownoutlier{q_\text{unk}}
\newcommand\dtrain{D_\text{train}}
\newcommand\dcalibrate{D_\text{calib}}
\newcommand\dtest{D_\text{test}}
\newcommand\conf{c}
\newcommand\fsrc{f_{\text{src}}}
\newcommand\fknown{f_{\text{src+known}}}

\title{Selective Question Answering under Domain Shift}

\author{Amita Kamath \hspace{10mm} Robin Jia \hspace{10mm} Percy Liang \\
  Computer Science Department, Stanford University \\
  \texttt{\{kamatha, robinjia, pliang\}@cs.stanford.edu} 
  \\}

\date{}

\begin{document}
\maketitle
\begin{abstract}
  To avoid giving wrong answers,
question answering (QA) models need to know when to abstain from answering.
Moreover,
users often ask questions that diverge from the model's training data,
making errors more likely and thus abstention more critical.
In this work, we propose the setting of selective question answering under domain shift,
in which a QA model is tested on a mixture of in-domain and out-of-domain data,
and must answer (i.e., not abstain on) as many questions as possible while maintaining high accuracy.
Abstention policies based solely on the model's softmax probabilities fare poorly,
since models are overconfident on out-of-domain inputs.
Instead, we train a calibrator to identify inputs on which the QA model errs,
and abstain when it predicts an error is likely.
Crucially, the calibrator benefits from observing the model's behavior on out-of-domain data,
even if from a different domain than the test data.
We combine this method with a SQuAD-trained QA model and evaluate on mixtures of SQuAD and five other QA datasets.
Our method answers $56\%$ of questions while maintaining $80\%$ accuracy;
in contrast, directly using the model's probabilities only answers $48\%$ at $80\%$ accuracy.
\end{abstract}

\section{Introduction}
\label{sec:introduction}
Question answering (QA) models have achieved impressive performance 
when trained and tested on examples from the same dataset,
but tend to perform poorly on examples that are out-of-domain (OOD) \citep{jia2017adversarial,chen2017reading,yogatama2019learning,talmor2019generalization,fisch2019mrqa}.
Deployed QA systems in search engines and personal assistants
need to gracefully handle OOD inputs,
as users often ask questions that fall outside of the system's training distribution.
While the ideal system would correctly answer all OOD questions,
such perfection is not attainable given limited training data \citep{geiger2019posing}.
Instead, we aim for a more achievable yet still challenging goal: 
models should \emph{abstain} when they are likely to err, thus avoiding showing wrong answers to users.
This general goal motivates the setting of selective prediction, in which a model outputs both a prediction and a scalar confidence,
and abstains on inputs where its confidence is low \citep{elyaniv2010foundations,geifman2017selective}. 

\begin{figure}[t] 
\centering
\includegraphics[width=\columnwidth]{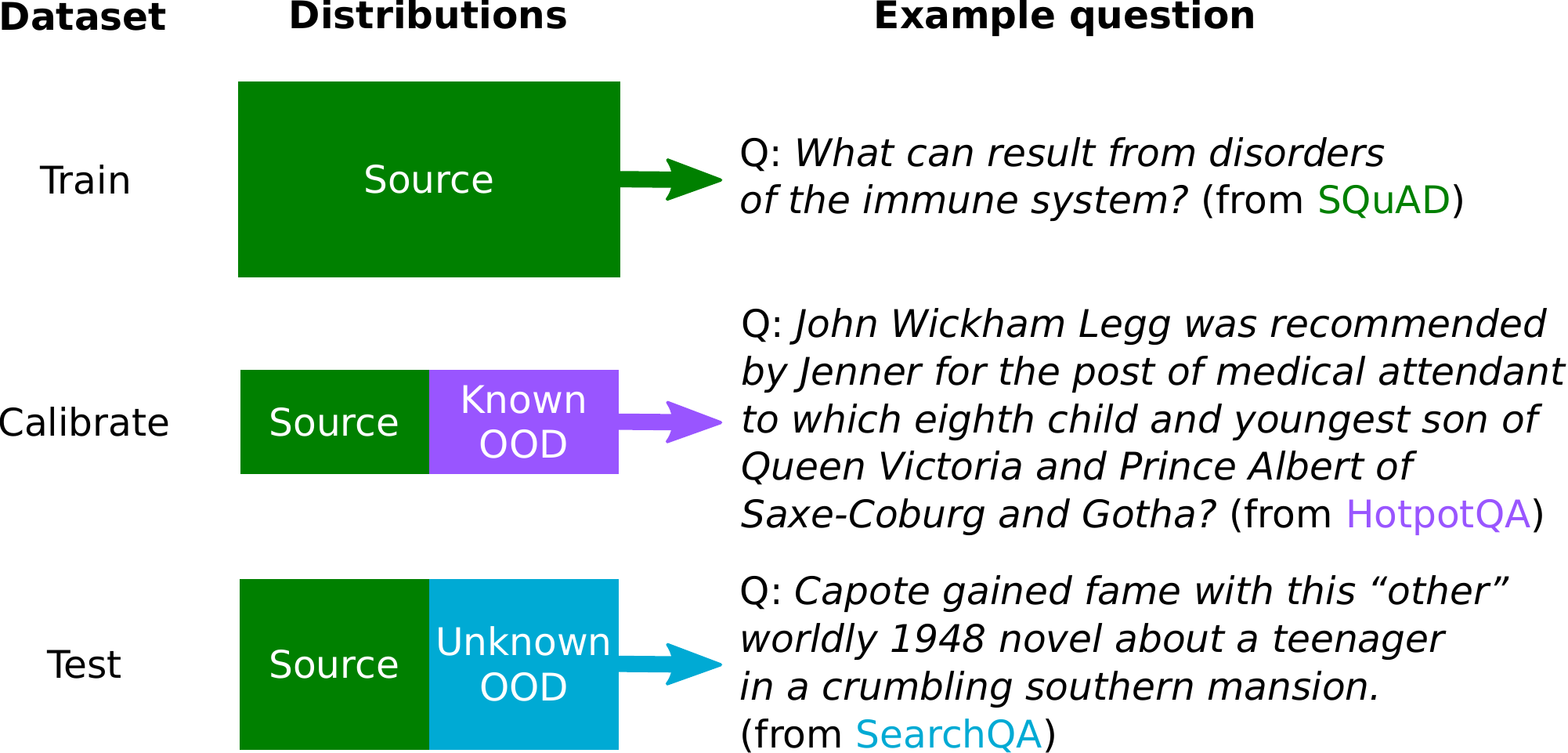}
\caption{
  Selective question answering under domain shift with a trained calibrator.
First, a QA model is trained only on source data.
Then, a calibrator is trained to predict whether the QA model was correct on any given example.
The calibrator's training data consists of both previously held-out source data and known OOD data.
Finally, the combined selective QA system is tested on a mixture of test data from the source distribution and an unknown OOD distribution.
}
\label{fig:distributions}
\end{figure}

In this paper, we propose the setting of \textbf{selective question answering under domain shift}, 
which captures two important aspects of real-world QA:
(i) test data often diverges from the training distribution,
and (ii) systems must know when to abstain.
We train a QA model on data from a \emph{source} distribution,
then evaluate selective prediction performance on a dataset that includes samples from both the source distribution and an \emph{unknown OOD} distribution.
This mixture simulates the likely scenario in which users only sometimes ask questions that are covered by the training distribution.
While the system developer knows nothing about the unknown OOD data,
we allow access to a small amount of data
from a third \emph{known OOD} distribution (e.g., OOD examples that they can foresee).

We first show that our setting is challenging because model softmax probabilities are unreliable estimates of confidence on out-of-domain data.
Prior work has shown that a strong baseline for in-domain selective prediction is MaxProb,
a method that abstains based on the probability assigned by the model to its highest probability prediction \citep{hendrycks2017baseline,lakshminarayanan2017simple}.
We find that MaxProb gives good confidence estimates on in-domain data,
but is overconfident on OOD data.
Therefore, MaxProb performs poorly in mixed settings:
it does not abstain enough on OOD examples, relative to in-domain examples.

We correct for MaxProb's overconfidence by using known OOD data to train a \emph{calibrator}---a classifier trained to predict whether the original QA model is correct or incorrect on a given example \citep{platt1999probabilistic,zadrozny2002transforming}.
While prior work in NLP trains a calibrator on in-domain data \citep{dong2018confidence},
we show this does not generalize to unknown OOD data as well as training on a mixture of in-domain and known OOD data.
\reffig{distributions} illustrates the problem setup and how the calibrator uses known OOD data.
We use a simple random forest calibrator over features derived
from the input example and the model's softmax outputs.

We conduct extensive experiments using SQuAD \citep{rajpurkar2016squad} as the source distribution and five other QA datasets as different OOD distributions.
We average across all 20 choices of using one as the unknown OOD dataset
and another as the known OOD dataset,
and test on a uniform mixture of SQuAD and unknown OOD data.
On average, the trained calibrator achieves $56.1\%$ coverage (i.e., the system answers $56.1\%$ of test questions)
while maintaining $80\%$ accuracy on answered questions,
outperforming MaxProb with the same QA model ($48.2\%$ coverage at $80\%$ accuracy),
using MaxProb and training the QA model on both SQuAD and the known OOD data ($51.8\%$ coverage),
and training the calibrator only on SQuAD data ($53.7\%$ coverage).

In summary, our contributions are as follows:  \\
\indent (1) We propose a novel setting, selective question answering under domain shift, 
that captures the practical necessity of knowing when to abstain on test data that differs from the training data. \\
\indent (2) We show that QA models are overconfident on out-of-domain examples relative to in-domain examples,
which causes MaxProb to perform poorly in our setting. \\
\indent (3) We show that out-of-domain data, even from a different distribution than the test data,
can improve selective prediction under domain shift when used to train a calibrator.
\section{Related Work}
\label{sec:related_work}
Our setting combines extrapolation to out-of-domain data 
with selective prediction.
We also distinguish our setting from the tasks of identifying unanswerable questions and outlier detection.

\subsection{Extrapolation to out-of-domain data}
Extrapolating from training data to test data from a different distribution
is an important challenge for current NLP models \citep{yogatama2019learning}.
Models trained on many domains may still struggle to generalize to new domains,
as these may involve new types of questions or require different reasoning skills \citep{talmor2019generalization,fisch2019mrqa}.
Related work on domain adaptation also tries to generalize to new distributions,
but assumes some knowledge about the test distribution, such as unlabeled examples or a few labeled examples \citep{blitzer2006domain,daume07easyadapt};
we assume no such access to the test distribution, but instead make the weaker assumption of access to samples from a different OOD distribution.

\subsection{Selective prediction}
Selective prediction, in which a model can either predict or abstain on each test example, is a longstanding research area in machine learning \citep{chow1957optimum,elyaniv2010foundations,geifman2017selective}.
In NLP, \citet{dong2018confidence} use a calibrator to obtain better confidence estimates for semantic parsing.
\citet{rodriguez2019quizbowl} use a similar approach to decide when to answer QuizBowl questions.
These works focus on training and testing models on the same distribution,
whereas our training and test distributions differ.

\paragraph{Selective prediction under domain shift.}
Other fields have recognized the importance of selective prediction under domain shift. 
In medical applications, models may be trained and tested on different groups of patients, so selective prediction is needed to avoid costly errors \citep{feng2019selective}.
In computational chemistry, \citet{toplak2014assessment} use selective prediction techniques to estimate the set of (possibly out-of-domain) molecules for which a reactivity classifier is reliable.
To the best of our knowledge, our work is the first to study selective prediction under domain shift in NLP.

\paragraph{Answer validation.}
Traditional pipelined systems for open-domain QA often have dedicated systems for answer validation---judging whether a proposed answer is correct.
These systems often rely on external knowledge about entities \citep{magnini2002right,ko2007probabilistic}.
Knowing when to abstain has been part of past QA shared tasks like
RespubliQA \citep{penas2009respubliqa} and QA4MRE \citep{penas2013mre}.
IBM's Watson system for Jeopardy also uses
a pipelined approach for answer validation \citep{gondek2012framework}.
Our work differs by focusing on modern neural QA systems trained end-to-end, rather than pipelined systems,
and by viewing the problem of abstention in QA through the lens of selective prediction.

\subsection{Related goals and tasks}
\label{sec:related_but_distinct}
\paragraph{Calibration.}
Knowing when to abstain is closely related to calibration---having a model's output probability align with the true probability of its prediction \citep{platt1999probabilistic}.
A key distinction is that selective prediction metrics generally depend only on \emph{relative} confidences---systems are judged on their ability to rank correct predictions higher than incorrect predictions \citep{elyaniv2010foundations}.
In contrast, calibration error depends on the absolute confidence scores.
Nonetheless, we will find it useful to analyze calibration in \refsec{overconfidence}, as miscalibration on some examples but not others does imply poor relative ordering, and therefore poor selective prediction.
\citet{ovadia2019uncertainty} observe increases in calibration error under domain shift. 

\paragraph{Identifying unanswerable questions.}
In SQuAD 2.0, models must recognize when a paragraph does not entail an answer to a question \citep{rajpurkar2018squadrun}.
Sentence selection systems must rank passages that answer a question higher than passages that do not
\citep{wang2007qa,yang2015wikiqa}.
In these cases, the goal is to ``abstain'' when \emph{no} system (or person)
could infer an answer to the given question using the given passage.
In contrast, in selective prediction, the model should abstain when \emph{it}
would give a wrong answer if forced to make a prediction.

\paragraph{Outlier detection.}
We distinguish selective prediction under domain shift from outlier detection,
the task of detecting out-of-domain examples \citep{scholkopf1999support,hendrycks2017baseline,liang2018enhancing}.
While one could use an outlier detector for selective classification (e.g., by abstaining on all examples flagged as outliers), this would be too conservative,
as QA models can often get a non-trivial fraction of OOD examples correct  \citep{talmor2019generalization,fisch2019mrqa}.
\citet{hendrycks2019anomaly} use known OOD data for outlier detection by training models to have high entropy on OOD examples; in contrast, our setting rewards models for predicting correctly on OOD examples, not merely having high entropy.

\section{Problem Setup}
\label{sec:problem_setup}
We formally define the setting of selective prediction under domain shift,
starting with some notation for selective prediction in general. 

\subsection{Selective Prediction}
Given an input $x$, the selective prediction task is to output $(\hat{y}, \conf)$ where $\hat{y} \in Y(x)$, the set of answer candidates, and $\conf \in \mathbb{R}$ denotes the model's confidence. 
Given a threshold $\gamma \in \mathbb{R}$, the overall system predicts $\hat{y}$ if $\conf \geq \gamma$ and abstain otherwise.

The risk-coverage curve provides a standard way to evaluate selective prediction methods \citep{elyaniv2010foundations}.
For a test dataset $\dtest$, any choice of $\gamma$ has an associated \emph{coverage}---the fraction of $\dtest$ the model makes a prediction on---and \emph{risk}---the error on that fraction of $\dtest$.
As $\gamma$ decreases, coverage increases, but risk will usually also increase.
We plot risk versus coverage and evaluate on the area under this curve (AUC),
as well as the maximum possible coverage for a desired risk level. 
The former metric averages over all $\gamma$, painting an overall picture of selective prediction performance, while the latter evaluates at a particular choice of $\gamma$ corresponding to a specific level of risk tolerance.

\subsection{Selective Prediction under Domain Shift}
We deviate from prior work by considering the setting
where the model's training data $\dtrain$ and test data $\dtest$ are drawn from different distributions.
As our experiments demonstrate, this setting is challenging because standard QA models are overconfident on out-of-domain inputs.

To formally define our setting, we specify three data distributions. 
First, $\psource$ is the source distribution, from which a large training dataset $\dtrain$ is sampled.
Second, $\punknownoutlier$ is an \emph{unknown OOD distribution}, representing out-of-domain data encountered at test time.
The test dataset $\dtest$ is sampled from $p_{\text{test}}$, a mixture of $\psource$ and $\punknownoutlier$:
\begin{align}
p_{\text{test}} = \alpha \psource + (1-\alpha) \punknownoutlier
\end{align}
for $\alpha \in (0, 1)$.
We choose $\alpha=\frac12$, and examine the effect of changing this ratio in Section \ref{subsec:change_ratio}. 
Third, $\pknownoutlier$ is a \emph{known OOD distribution}, representing examples not in $\psource$ but 
from which the system developer has a small dataset $\dcalibrate$.

\subsection{Selective Question Answering}
While our framework is general, we focus on extractive question answering, as exemplified by SQuAD \citep{rajpurkar2016squad}, due to its practical importance
and the diverse array of available QA datasets in the same format.
The input $x$ is a passage-question pair $(p, q)$,
and the set of answer candidates $Y(x)$ is all spans of the passage $p$.
A \emph{base model} $f$ defines a probability distribution $f(y \mid x)$ over $Y(x)$.
All selective prediction methods we consider choose $\hat{y} = \arg\max_{y' \in Y(x)} f(y' \mid x)$,
but differ in their associated confidence $c$.

\section{Methods}
\label{sec:approach}
Recall that our setting differs from the standard selective prediction setting in two ways: unknown OOD data drawn from $\punknownoutlier$ appears at test time, and known OOD data drawn from $\pknownoutlier$ is available to the system. 
Intuitively, we expect that systems must use the known OOD data to generalize to the unknown OOD data.
In this section, we present three standard selective prediction methods for in-domain data, and show how they can be adapted to use data from $\pknownoutlier$.

\subsection{MaxProb}
\label{subsec:maxprob}
The first method, MaxProb, directly uses the probability assigned by the base model to $\hat{y}$ as an estimate of confidence.
Formally, MaxProb with model $f$ estimates confidence on input $x$ as:
\begin{align}
c_{\text{MaxProb}} = f(\hat{y} \mid x) = \max_{y' \in Y(x)} f(y' \mid x). 
\end{align}

MaxProb is a strong baseline for our setting. 
Across many tasks, MaxProb has been shown to distinguish
in-domain test examples that the model gets right from ones the model gets wrong \citep{hendrycks2017baseline}.
MaxProb is also a strong baseline for outlier detection,
as it is lower for out-of-domain examples than in-domain examples \citep{lakshminarayanan2017simple,liang2018enhancing,hendrycks2019anomaly}. 
This is desirable for our setting:
models make more mistakes on OOD examples, so they should abstain more on OOD examples than in-domain examples.

MaxProb can be used with any base model $f$.
We consider two such choices:
a model $\fsrc$ trained only on $\dtrain$, or a model $\fknown$ trained on the union of $\dtrain$ and $\dcalibrate$.

\subsection{Test-time Dropout}
\label{sec:dropout}
For neural networks, another standard approach to estimate confidence is to use dropout at test time.
\citet{gal2016dropout} showed that dropout gives good confidence estimates on OOD data.

Given an input $x$ and model $f$, we compute $f$ on $x$ with $K$ different dropout masks, obtaining prediction distributions $\hat{p}_1, \dotsc, \hat{p}_K$, where each $\hat{p}_i$ is a probability distribution over $Y(x)$.
We consider two statistics of these $\hat{p}_i$'s that are commonly used as confidence estimates.
First, we take the mean of $\hat{p}_i(\hat{y})$ across all $i$ \citep{lakshminarayanan2017simple}:
\begin{align}
c_{\text{DropoutMean}} = \frac1K \sum_{i=1}^K \hat{p}_i(\hat{y}).
\end{align}
This can be viewed as ensembling the predictions across all $K$ dropout masks by averaging them.

Second, we take the negative variance of the $\hat{p}_i(\hat{y})$'s \citep{feinman2017detecting,smith2018understanding}:
\begin{align}
c_{\text{DropoutVar}} = -\mathrm{Var}[\hat{p}_1(\hat{y}), \dotsc, \hat{p}_K(\hat{y})].
\end{align}
Higher variance corresponds to greater uncertainty, and hence favors abstaining.
Like MaxProb, dropout can be used either with $f$ trained only on $\dtrain$, or on both $\dtrain$ and the known OOD data.

Test-time dropout has practical disadvantages compared to MaxProb.
It requires access to internal model representations, whereas MaxProb only requires black box access to the base model (e.g., API calls to a trained model).
Dropout also requires $K$ forward passes of the base model, leading to a $K$-fold increase in runtime.

\subsection{Training a calibrator}
Our final method trains a calibrator to predict when a base model (trained only on data from $\psource$) is correct
\citep{platt1999probabilistic,dong2018confidence}.
We differ from prior work by training the calibrator
on a mixture of data from $\psource$ and $\pknownoutlier$,
anticipating the test-time mixture of $\psource$ and $\punknownoutlier$.
More specifically, 
we hold out a small number of $\psource$ examples from base model training,
and train the calibrator on the union of these examples and the $\pknownoutlier$ examples.
We define $c_{\text{Calibrator}}$ to be the prediction probability of the calibrator.

The calibrator itself could be any binary classification model.
We use a random forest classifier with seven features:
passage length, the length of the predicted answer $\hat{y}$, and the top five softmax probabilities output by the model. 
These features require only a minimal amount of domain knowledge to define.
\citet{rodriguez2019quizbowl} similarly used multiple softmax probabilities to decide when to answer questions.
The simplicity of this model makes the calibrator fast to train when given new data from $\pknownoutlier$,
especially compared to re-training the QA model on that data.

We experiment with four variants of the calibrator.
First, to measure the impact of using known OOD data, we change the calibrator's training data: it can be trained either on data from $\psource$ only, or both $\psource$ and $\pknownoutlier$ data as described. 
Second, we consider a modification where instead of the model's probabilities, we use probabilities from the mean ensemble over dropout masks, as described in \refsec{dropout}, and also add $c_{\text{DropoutVar}}$ as a feature.
As discussed above, dropout features are costly to compute and assume white-box access to the model, but may result in better confidence estimates. 
Both of these variables can be changed independently, leading to four configurations.

\section{Experiments and Analysis}
\label{sec:experiments}
\subsection{Experimental Details}
\label{subsec:experimental_details}
\paragraph{Data.}
We use SQuAD 1.1 \citep{rajpurkar2016squad} as the source dataset and
five other datasets as OOD datasets:
NewsQA \citep{trischler2017newsqa}, TriviaQA \citep{joshi2017triviaqa}, SearchQA \citep{dunn2017searchqa}, HotpotQA \citep{yang2018hotpotqa}, and Natural Questions \citep{kwiatkowski2019natural}.\footnote{We consider these different datasets to represent different domains, hence our usage of the term ``domain shift.''}
These are all extractive question answering datasets where all questions are answerable; 
however, they vary widely in the nature of passages (e.g., Wikipedia, news, web snippets),
questions (e.g., Jeopardy and trivia questions), 
and relationship between passages and questions (e.g., whether questions are written based on passages, or passages retrieved based on questions).
We used the preprocessed data from the MRQA 2019 shared task \citep{fisch2019mrqa}. 
For HotpotQA, we focused on multi-hop questions by selecting only ``hard'' examples, as defined by \citet{yang2018hotpotqa}.
In each experiment, two different OOD datasets are chosen as $\pknownoutlier$ and $\punknownoutlier$. 
All results are averaged over all 20 such combinations, unless otherwise specified.
We sample 2,000 examples from $\pknownoutlier$ for $\dcalibrate$,
and 4,000 SQuAD and 4,000 $\punknownoutlier$ examples for $\dtest$.
We evaluate using exact match (EM) accuracy, as defined by SQuAD \citep{rajpurkar2016squad}.
Additional details can be found in \refapp{dataset_sources}.

\paragraph{QA model.}
For our QA model, we use the BERT-base SQuAD 1.1 model trained for 2 epochs \citep{devlin2019bert}.
We train six models total: one $\fsrc$ and five $\fknown$'s, one for each OOD dataset.

\paragraph{Selective prediction methods.}
For test-time dropout, we use $K=30$ different dropout masks, as in \citet{dong2018confidence}.
For our calibrator, we use the random forest implementation from Scikit-learn \citep{pedregosa2011sklearn}.
We train on 1,600 SQuAD examples and 1,600 known OOD examples, and use the remaining 400 SQuAD and 400 known OOD examples as a validation set to tune calibrator hyperparameters via grid search.
We average our results over 10 random splits of this data.
When training the calibrator only on $\psource$, we use 3,200 SQuAD examples for training and 800 for validation,
to ensure equal dataset sizes.
Additional details can be found in \refapp{calibrator_features_model}.

\subsection{Main results}
\label{sec:main_results}
\paragraph{Training a calibrator with $\pknownoutlier$ outperforms other methods.}
\reftab{main_table} compares all methods that do not use test-time dropout.
Compared to MaxProb with $\fknown$, the calibrator has $4.3$ points and $6.7$ points higher coverage at $80\%$ and $90\%$ accuracy respectively, and $1.1$ points lower AUC.\footnote{$95\%$ confidence interval is $[1.01, 1.69]$, using the paired bootstrap test with 1000 bootstrap samples.}
This demonstrates that training a calibrator is a better use of known OOD data than training a QA model.
The calibrator trained on both $\psource$ and $\pknownoutlier$ also outperforms the calibrator trained on $\psource$ alone by $2.4\%$ coverage at $80\%$ accuracy.
All methods perform far worse than the optimal selective predictor with the given base model,
though achieving this bound may not be realistic.\footnote{As the QA model has fixed accuracy $< 100\%$ on $\dtest$, it is impossible to achieve $0\%$ risk at $100\%$ coverage.}

\paragraph{Test-time dropout improves results but is expensive.}
\reftab{main_table_dropout} shows results for methods that use test-time dropout, as described in \refsec{dropout}. 
The negative variance of $\hat{p}_i(\hat{y})$'s across dropout masks serves poorly as an estimate of confidence, but the mean performs well. The best performance is attained by the calibrator using dropout features,
which has $3.9\%$ higher coverage at $80\%$ accuracy than the calibrator with non-dropout features.
Since test-time dropout introduces substantial (i.e., $K$-fold) runtime overhead,
our remaining analyses focus on methods without test-time dropout.

\begin{table}[t]
\resizebox{\columnwidth}{!}{%
\begin{tabular}{lccc}
\toprule
   & \textbf{\begin{tabular}[c]{@{}c@{}}AUC\\ $\downarrow$\end{tabular}} & \textbf{\begin{tabular}[c]{@{}c@{}}Cov @\\ Acc=80\%\\ $\uparrow$\end{tabular}} & \textbf{\begin{tabular}[c]{@{}c@{}}Cov @\\ Acc=90\%\\ $\uparrow$\end{tabular}} \\ \midrule
\begin{tabular}[c]{@{}l@{}}\textbf{Train QA model on SQuAD}\\ MaxProb\\ Calibrator ($\psource$ only)\\Calibrator ($\psource$ and $\pknownoutlier$)\\ Best possible\end{tabular}       & \begin{tabular}[c]{@{}c@{}}\\ 20.54\\ 19.27\\ \textbf{18.47}\\ 9.64\end{tabular}                 & \begin{tabular}[c]{@{}c@{}}\\ 48.23\\ 53.67\\\textbf{56.06}\\ 74.92\end{tabular}                                   & \begin{tabular}[c]{@{}c@{}}\\ 21.07\\ 26.68\\\textbf{29.42}\\ 66.59\end{tabular}                                   \\ \hline
\begin{tabular}[c]{@{}l@{}}\textbf{Train QA model on SQuAD +}\\ \textbf{known OOD}\\ MaxProb\\ Best possible\end{tabular} & \begin{tabular}[c]{@{}c@{}}\\ \\ 19.61\\ 8.83\end{tabular}                     & \begin{tabular}[c]{@{}c@{}}\\ \\ 51.75\\ 76.80\end{tabular}                                       & \begin{tabular}[c]{@{}c@{}}\\ \\ 22.76\\ 68.26\end{tabular}                                       \\ \bottomrule
\end{tabular}}
  \caption{\label{tab:main_table} Results for methods without test-time dropout. The calibrator with access to $\pknownoutlier$ outperforms all other methods.
$\downarrow$: lower is better.  $\uparrow$: higher is better. 
  }
\end{table}

\begin{table}[t]
\resizebox{\columnwidth}{!}{%
\begin{tabular}{lccc}
\toprule
   & \textbf{\begin{tabular}[c]{@{}c@{}}AUC\\ $\downarrow$\end{tabular}} & \textbf{\begin{tabular}[c]{@{}c@{}}Cov @\\ Acc=80\%\\ $\uparrow$\end{tabular}} & \textbf{\begin{tabular}[c]{@{}c@{}}Cov @\\ Acc=90\%\\ $\uparrow$\end{tabular}} \\ \midrule
\begin{tabular}[c]{@{}l@{}}\textbf{Train QA model on SQuAD}\\ Test-time dropout (--var) \\ Test-time dropout (mean)\\ Calibrator ($\psource$ only)\\Calibrator ($\psource$ and $\pknownoutlier$)\\ Best possible\end{tabular}       & \begin{tabular}[c]{@{}c@{}}\\ 28.13\\18.35\\ 17.84\\\textbf{17.31}\\ 9.64\end{tabular}                 & \begin{tabular}[c]{@{}c@{}}\\ 24.50\\ 57.49\\ 58.35\\\textbf{59.99}\\ 74.92\end{tabular}                                   & \begin{tabular}[c]{@{}c@{}}\\ 15.40 \\29.55\\ 34.27\\\textbf{34.99}\\ 66.59\end{tabular}                                   \\ \hline
\begin{tabular}[c]{@{}l@{}}\textbf{Train QA model on SQuAD +}\\ \textbf{known OOD}\\ Test-time dropout (--var) \\Test-time dropout (mean)\\ Best possible\end{tabular} & \begin{tabular}[c]{@{}c@{}}\\ \\ 26.67 \\17.72\\ 8.83\end{tabular}                     & \begin{tabular}[c]{@{}c@{}}\\ \\ 26.74\\59.60\\ 76.80\end{tabular}                                       & \begin{tabular}[c]{@{}c@{}}\\ \\ 15.95\\30.40\\ 68.26\end{tabular}                                       \\ \bottomrule
\end{tabular}}
\caption{\label{tab:main_table_dropout} Results for methods that use test-time dropout. Here again, the calibrator with access to $\pknownoutlier$ outperforms all other methods.
  }
\end{table}

\paragraph{The QA model has lower non-trivial accuracy on OOD data.}
Next, we motivate our focus on selective prediction, as opposed to outlier detection, by showing that
the QA model still gets a non-trivial fraction of OOD examples correct.
\reftab{em_table} shows the (non-selective) exact match scores for all six QA models used in our experiments on all datasets. 
All models get around $80\%$ accuracy on SQuAD, and around $40\%$ to $50\%$ accuracy on most OOD datasets.
Since OOD accuracies are much higher than $0\%$, abstaining on all OOD examples would be overly conservative.\footnote{In \refsec{outlier_detection}, we confirm that an outlier detector does not achieve good selective prediction performance.}
At the same time, since OOD accuracy is worse than in-domain accuracy, a good selective predictor should answer more in-domain examples and fewer OOD examples.
Training on 2,000 $\pknownoutlier$ examples 
does not significantly help the base model extrapolate to other $\punknownoutlier$ distributions.

\begin{table*}[t]
\centering
\resizebox{0.7\textwidth}{!}{%
\begin{tabular}{lcccccc}
\toprule
\multicolumn{1}{c}{\textbf{Train Data $\downarrow$ / Test Data $\rightarrow$}} & \textbf{SQuAD} & \textbf{TriviaQA} & \textbf{HotpotQA} & \textbf{NewsQA} & \textbf{\begin{tabular}[c]{@{}c@{}}Natural\\ Questions\end{tabular}} & \textbf{SearchQA} \\ \midrule
\textbf{SQuAD only}                                 & 80.95          & 48.43             & 44.88             & 40.45           & 42.78                                                                & 17.98             \\ 
\textbf{SQuAD + 2K TriviaQA}                        & 81.48          & \color{gray}(50.50)             & 43.95             & 39.15           & 47.05                                                                & 25.23             \\ 
\textbf{SQuAD + 2K HotpotQA}                        & 81.15          & 49.35             & \color{gray}(53.60)             & 39.85           & 48.18                                                                & 24.40             \\ 
\textbf{SQuAD + 2K NewsQA}                          & 81.50          & 50.18             & 42.88             & \color{gray}(44.00)           & 47.08                                                                & 20.40             \\ 
\textbf{SQuAD + 2K NaturalQuestions}                & 81.48          & 51.43             & 44.38             & 40.90           & \color{gray}(54.85)                                                                & 25.95             \\ 
\textbf{SQuAD + 2K SearchQA}                        & 81.60          & 56.58             & 44.30             & 40.15           & 47.05                                                                & \color{gray}(59.80)             \\ \bottomrule
\end{tabular}
}
\caption{Exact match accuracy for all six QA models on all six test QA datasets.
Training on $\dcalibrate$ improves accuracy on data from the same dataset (diagonal), but generally does not improve accuracy on data from $\punknownoutlier$.}
\label{tab:em_table}
\end{table*}

\begin{figure}[t]
\centering
\includegraphics[width=0.8\columnwidth]{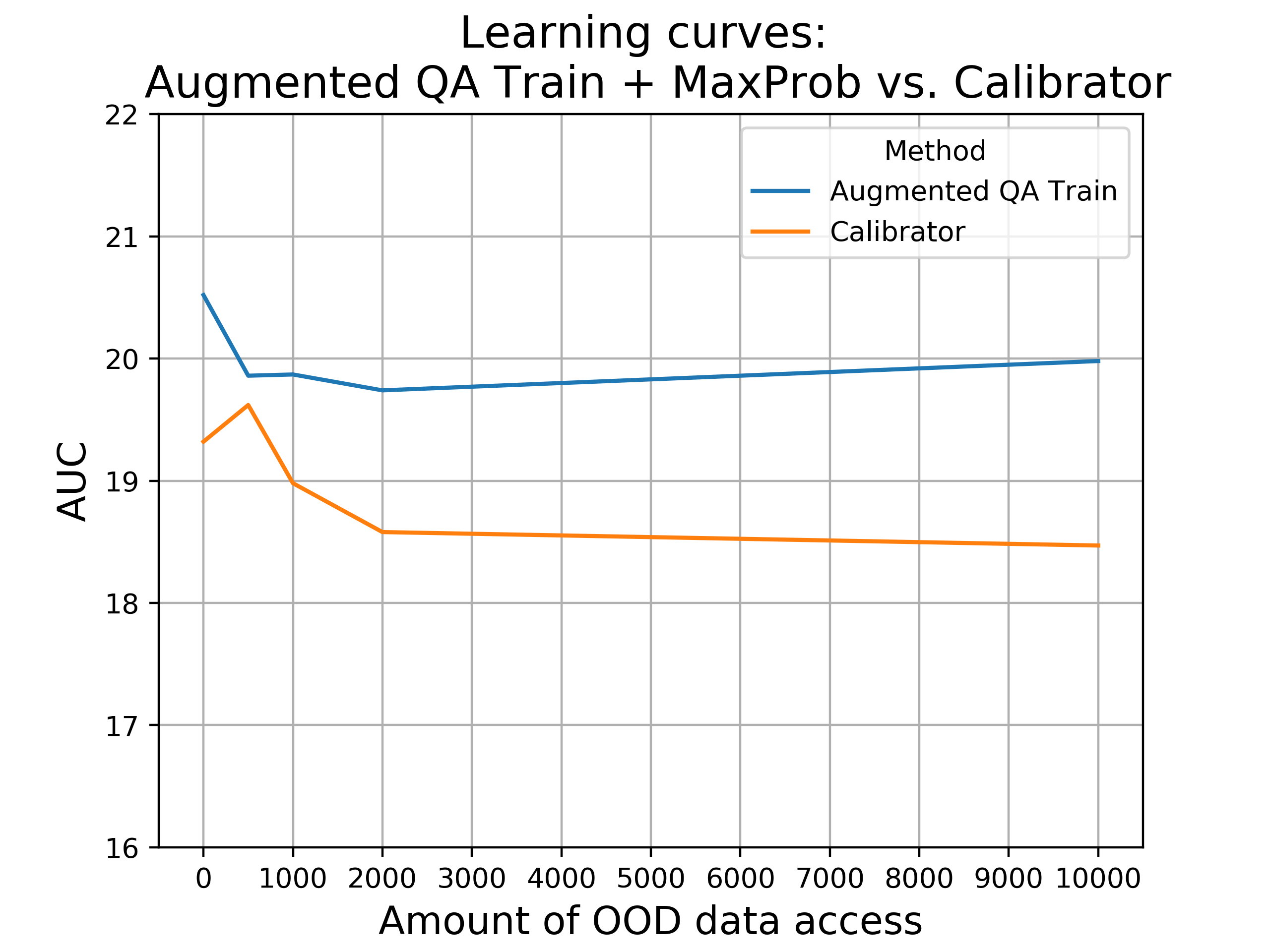}
\caption{Area under the risk-coverage curve as a function of how much data from $\pknownoutlier$ is available.
At all points, using data from $\pknownoutlier$ to train the calibrator is more effective than using it for QA model training.
}
\label{fig:learning_curve}
\end{figure}

\paragraph{Results hold across different amounts of known OOD data.}
As shown in \reffig{learning_curve}, across all amounts of known OOD data,
using it to train and validate the calibrator (in an 80--20 split) performs better than adding all of it to the QA training data and using MaxProb.

\subsection{Overconfidence of MaxProb}
\label{sec:overconfidence}
\begin{figure}[t]
    \centering
    \begin{subfigure}[t]{0.23\textwidth}
        \centering
        \includegraphics[width=\linewidth]{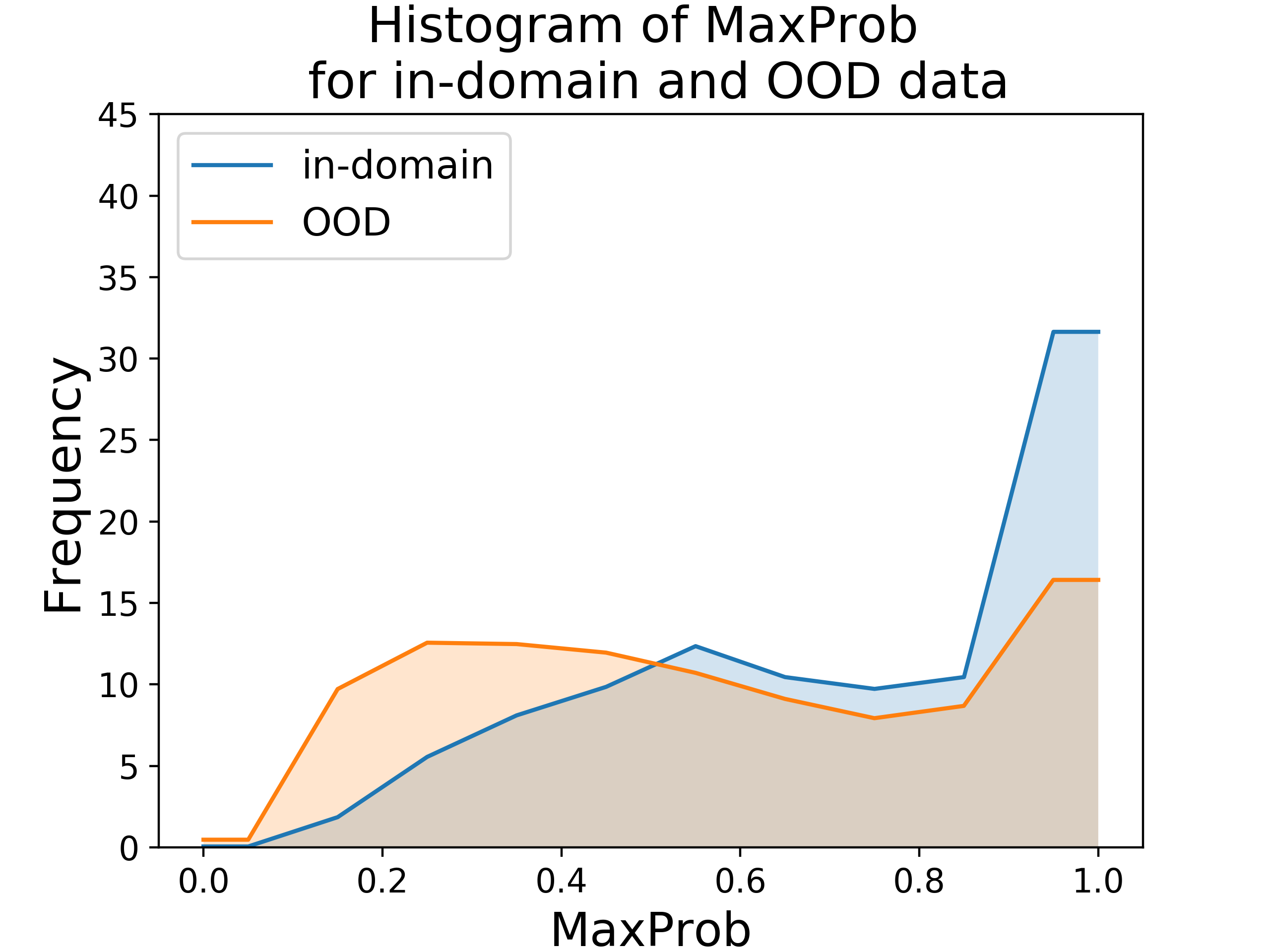} 
        \caption{} \label{fig:maxprob_density}
    \end{subfigure}
    \begin{subfigure}[t]{0.23\textwidth}
        \centering
        \includegraphics[width=\linewidth]{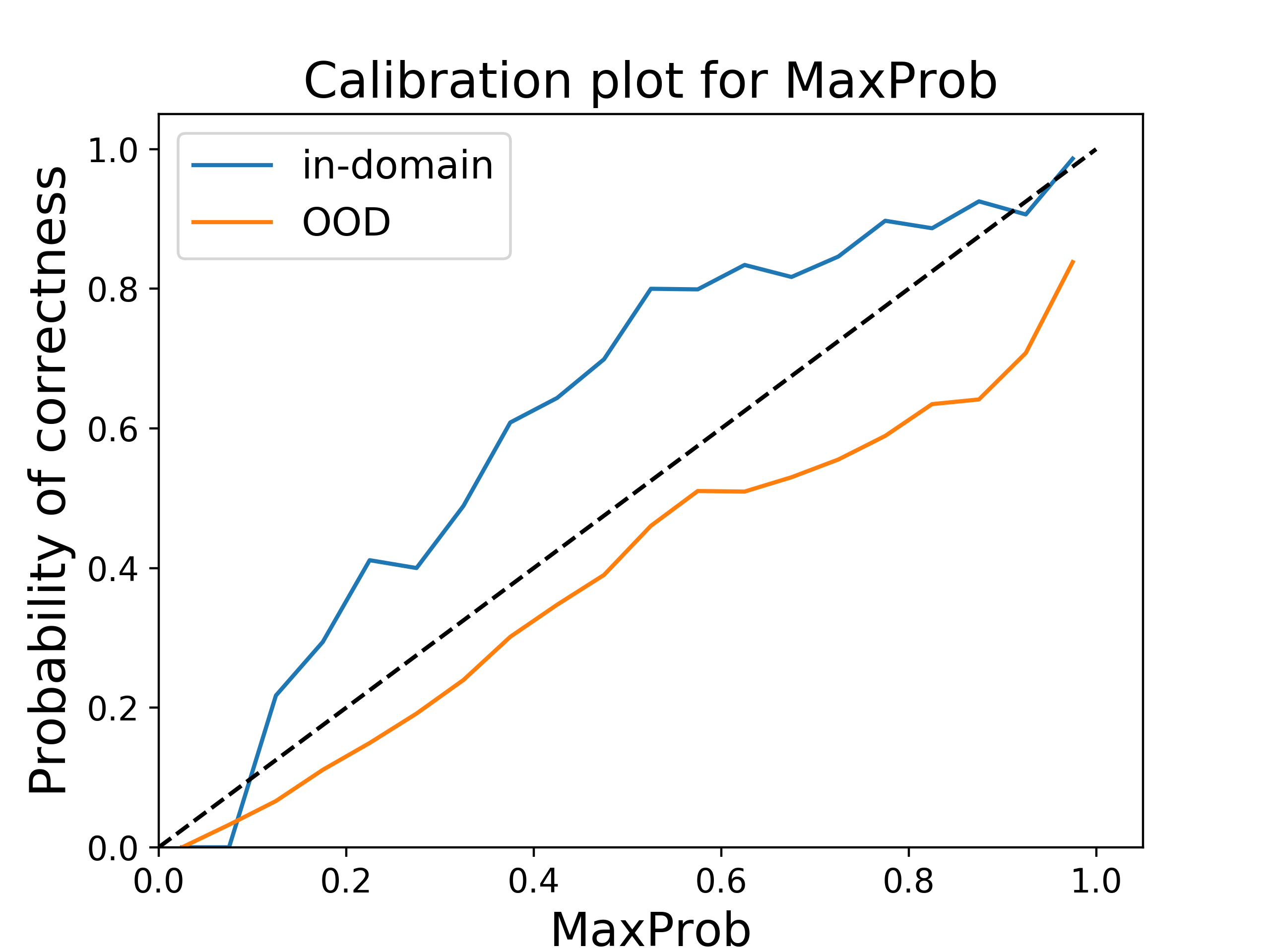} 
        \caption{} \label{fig:maxprob_calibration}
    \end{subfigure}
    \begin{subfigure}[t]{0.23\textwidth}
        \centering
        \includegraphics[width=\linewidth]{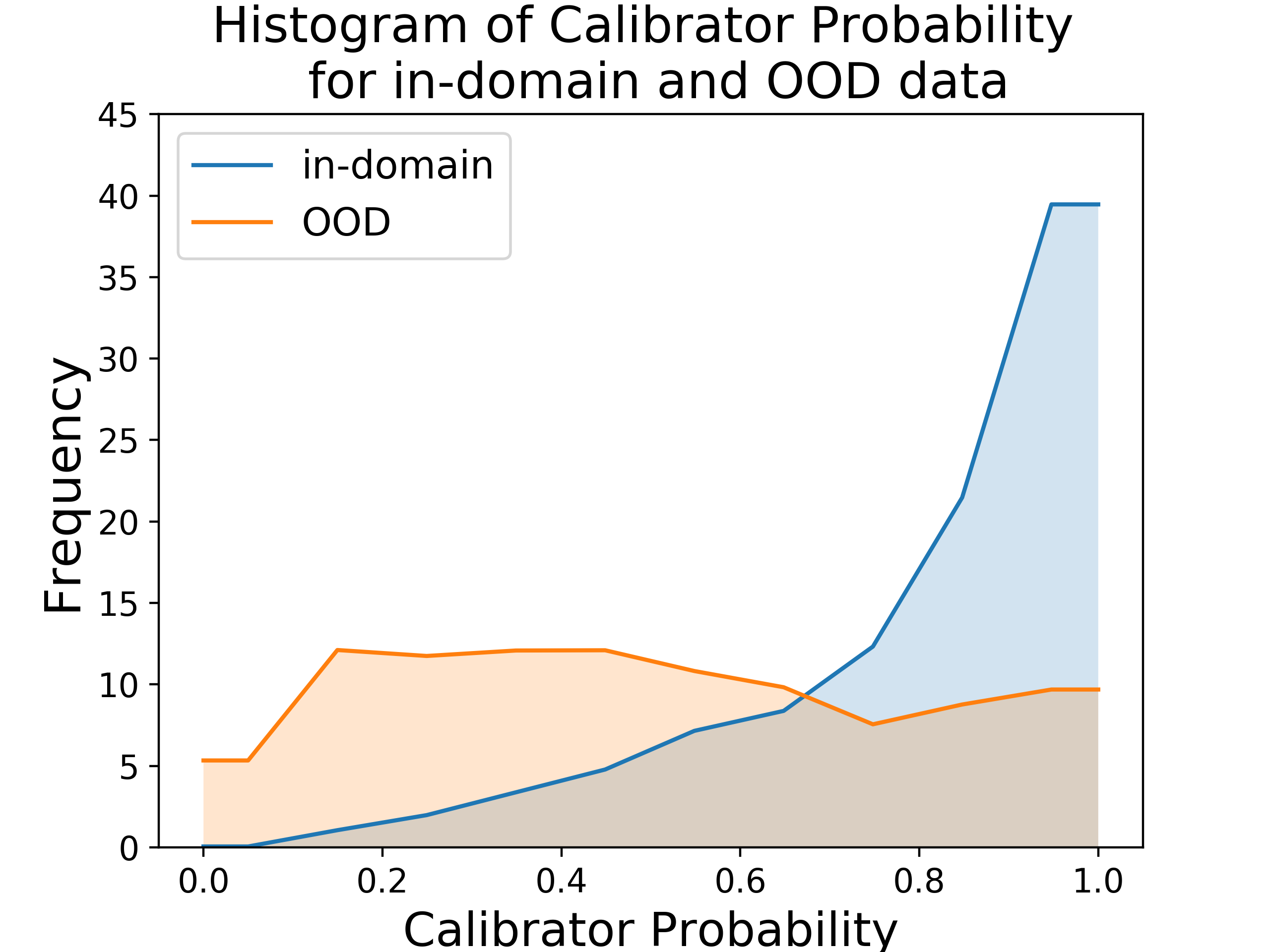} 
        \caption{} \label{fig:calibrator_density}
    \end{subfigure}
    \begin{subfigure}[t]{0.23\textwidth}
        \centering
        \includegraphics[width=\linewidth]{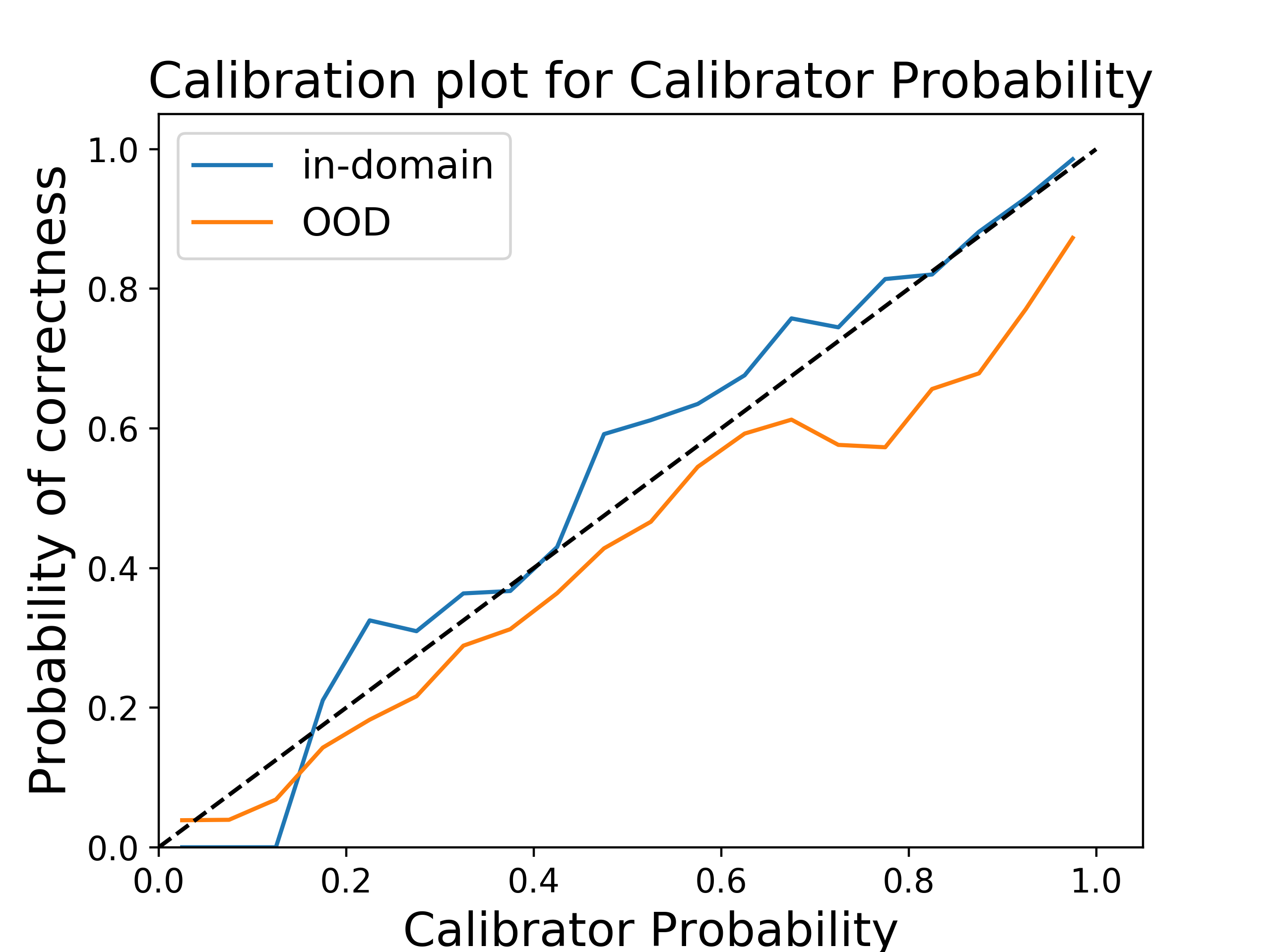} 
        \caption{} \label{fig:calibrator_calibration}
    \end{subfigure}
    \caption{
MaxProb is lower on average for OOD data than in-domain data
(\subref{fig:maxprob_density}),
but it is still overconfident on OOD data:
when plotting the true probability of correctness vs. MaxProb (\subref{fig:maxprob_calibration}),
the OOD curve is below the $y=x$ line, indicating MaxProb overestimates the probability that the prediction is correct.
The calibrator assigns lower confidence on OOD data (\subref{fig:calibrator_density})
and has a smaller gap between in-domain and OOD curves (\subref{fig:calibrator_calibration}), 
indicating improved calibration.
}
\end{figure}

We now show why MaxProb performs worse in our setting compared to the in-domain setting:
it is miscalibrated on out-of-domain examples. 
\reffig{maxprob_density} shows that MaxProb values are generally lower for OOD examples than in-domain examples,
following previously reported trends \citep{hendrycks2017baseline,liang2018enhancing}.
However, the MaxProb values are still too high out-of-domain.
Figure \ref{fig:maxprob_calibration} shows that MaxProb is not well calibrated: 
it is underconfident in-domain, and overconfident out-of-domain.\footnote{The in-domain underconfidence 
is because SQuAD (and some other datasets) provides only one answer at training time, but multiple answers are considered correct at test time. 
In \refapp{underconfidence_squad}, we show that removing multiple answers makes MaxProb well-calibrated in-domain; it stays overconfident out-of-domain.}
For example, for a MaxProb of $0.6$, the model is about $80\%$ likely to get the question correct if it came from SQuAD (in-domain), and $45\%$ likely to get the question correct if it was OOD. 
When in-domain and OOD examples are mixed at test time,
MaxProb therefore does not abstain enough on the OOD examples.
\reffig{calibrator_calibration} shows that the calibrator is better calibrated,
even though it is not trained on any unknown OOD data.
In \refapp{per_domain_breakdown}, we show that the calibrator abstains on more OOD examples than MaxProb.

Our finding that the BERT QA model is not overconfident in-domain aligns with \citet{hendrycks2019pretraining}, who found that pre-trained computer vision models are better calibrated than models trained from scratch, as pre-trained models can be trained for fewer epochs. Our QA model is only trained for two epochs, as is standard for BERT.
Our findings also align with \citet{ovadia2019uncertainty}, who find that computer vision and text classification models are poorly calibrated out-of-domain even when well-calibrated in-domain.
Note that miscalibration out-of-domain does not imply poor selective prediction on OOD data,
but does imply poor selective prediction in our mixture setting.

\subsection{Extrapolation between datasets}
\begin{figure}[t]
{
\centering
\includegraphics[width=\columnwidth]{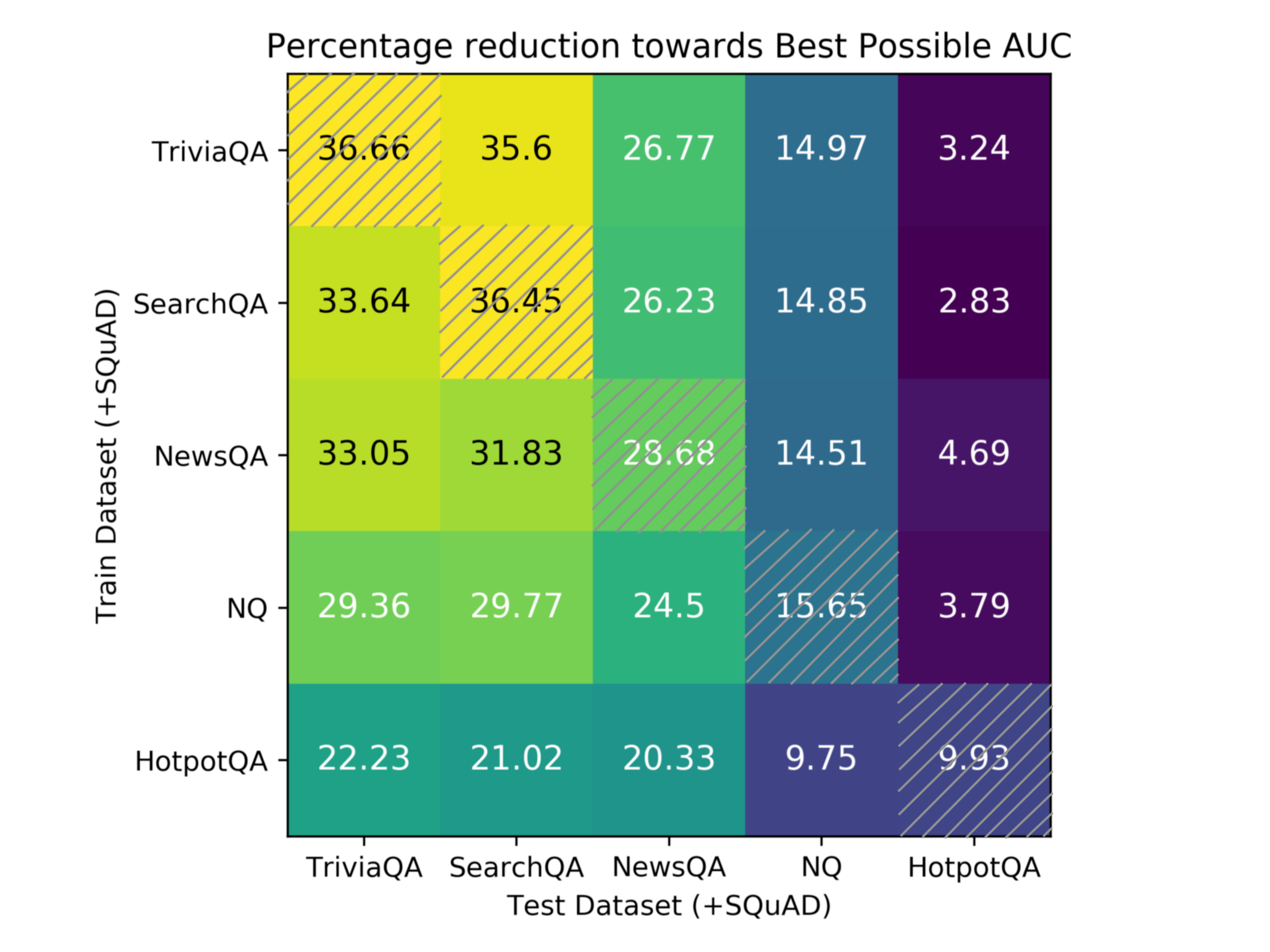}
\caption{
Results for different choices of $\pknownoutlier$ (y-axis) and $\punknownoutlier$ (x-axis).
For each pair, we report the percent AUC improvement of the trained calibrator over MaxProb,
relative to the total possible improvement.
Datasets that use similar passages (e.g., SearchQA and TriviaQA) help each other the most. 
Main diagonal elements (shaded) assume access to $\punknownoutlier$ (see \refsec{oracle}).
}
\label{fig:extrapolation_heatmap}
}
\end{figure}

We next investigated how choice of $\pknownoutlier$ affects generalization of the calibrator to $\punknownoutlier$.
\reffig{extrapolation_heatmap} shows the percentage reduction between MaxProb and optimal AUC achieved by the trained calibrator.
The calibrator outperforms MaxProb over all dataset combinations,
with larger gains when $\pknownoutlier$ and $\punknownoutlier$ are similar.
For example, samples from TriviaQA help generalization to SearchQA and vice versa; both use web snippets as passages.
Samples from NewsQA, the only other non-Wikipedia dataset, are also helpful for both.
On the other hand, no other dataset significantly helps generalization to HotpotQA, likely due to HotpotQA's unique focus on multi-hop questions.

\subsection{Calibrator feature ablations}
We determine the importance of each feature of the calibrator by removing each of its features individually, leaving the rest. 
From Table \ref{tab:ablations}, we see that the most important features are the softmax probabilities and the passage length.
Intuitively, passage length is meaningful both because longer passages have more answer candidates,
and because passage length differs greatly between different domains.

\begin{table}[t]
\resizebox{\columnwidth}{!}{%
\begin{tabular}{lccc}
\toprule
   & \textbf{\begin{tabular}[c]{@{}c@{}}AUC\\ $\downarrow$\end{tabular}} & \textbf{\begin{tabular}[c]{@{}c@{}}Cov @\\ Acc=80\%\\ $\uparrow$\end{tabular}} & \textbf{\begin{tabular}[c]{@{}c@{}}Cov @\\ Acc=90\%\\ $\uparrow$\end{tabular}} \\ \midrule
\begin{tabular}[c]{@{}l@{}}\textbf{All features}\\ --Top softmax probability\\ --2nd:5th highest \\ softmax probabilities\\--All softmax probabilities\\ --Context length \\ --Prediction length\end{tabular}       & \begin{tabular}[c]{@{}c@{}}\textbf{18.47}\\ 18.61\\ 19.11\\ \\26.41\\ 19.79\\ 18.6\end{tabular}                 & \begin{tabular}[c]{@{}c@{}}\textbf{56.06}\\ 55.46\\ 54.29\\ \\24.57\\ 51.73\\ 55.67\end{tabular}                                   & \begin{tabular}[c]{@{}c@{}}\textbf{29.42}\\ 29.27\\ 26.67\\ \\0.08\\ 24.24\\ 29.30\end{tabular}                                   \\ \bottomrule
\end{tabular}}
\caption{\label{tab:ablations}Performance of the calibrator as each of its features is removed individually, leaving the rest. The base model's softmax probabilities are important features, as is passage length.
  }
\end{table}

\subsection{Error analysis}
We examined calibrator errors on two pairs of $\pknownoutlier$ and $\punknownoutlier$---one similar pair of datasets and one dissimilar.
For each, we sampled 100 errors in which the system confidently gave a wrong answer (overconfident), and 100 errors in which the system abstained but would have gotten the question correct if it had answered (underconfident). These were sampled from the 1000 most overconfident or underconfident errors, respectively.

\paragraph{$\pknownoutlier=\text{NewsQA}$, $\punknownoutlier=\text{TriviaQA}$.}
These two datasets are from different non-Wikipedia sources.
$62\%$ of overconfidence errors are due to the model predicting valid alternate answers, or span mismatches---the model predicts a slightly different span than the gold span, and should be considered correct; thus the calibrator was not truly overconfident.
This points to the need to improve QA evaluation metrics \citep{chen2019evaluating}.
$45\%$ of underconfidence errors are due to the passage requiring coreference resolution over long distances, including with the article title. Neither SQuAD nor NewsQA passages have coreference chains as long or contain titles, so it is unsurprising that the calibrator struggles on these cases. 
Another $25\%$ of underconfidence errors were cases in which there was insufficient evidence in the paragraph to answer the question (as TriviaQA was constructed via distant supervision), so the calibrator was not incorrect to assign low confidence.
$16\%$ of all underconfidence errors also included phrases that would not be common in SQuAD and NewsQA, such as using \nl{said bye bye} for \nl{banned.}

\paragraph{$\pknownoutlier=\text{NewsQA}$, $\punknownoutlier=\text{HotpotQA}$.}
These two datasets are dissimilar from each other in multiple ways.
HotpotQA uses short Wikipedia passages and focuses on multi-hop questions;
NewsQA has much longer passages from news articles and does not focus on multi-hop questions.
$34\%$ of the overconfidence errors are due to valid alternate answers or span mismatches.
On $65\%$ of the underconfidence errors, the correct answer was the only span in the passage that could plausibly answer the question, suggesting that the model arrived at the answer due to artifacts in HotpotQA
that facilitate guesswork \citep{chen2019understanding,min2019compositional}. 
In these situations, the calibrator's lack of confidence is therefore justifiable.

\subsection{Relationship with Unanswerable Questions}
We now study the relationship between selective prediction and identifying unanswerable questions.
\paragraph{Unanswerable questions do not aid selective prediction.} 
We trained a QA model on SQuAD 2.0 \citep{rajpurkar2018squadrun}, which augments SQuAD 1.1 with unanswerable questions.
Our trained calibrator with this model gets $18.38$ AUC, which is very close to the $18.47$ for the model trained on SQuAD 1.1 alone.
MaxProb also performed similarly with the SQuAD 2.0 model ($20.81$ AUC) and SQuAD 1.1 model ($20.54$ AUC).
\paragraph{Selective prediction methods do not identify unanswerable questions.} 
For both MaxProb and our calibrator, we pick a threshold $\gamma' \in \mathbb{R}$ and predict that a question is unanswerable if the confidence $c < \gamma'$.
We choose $\gamma'$ to maximize SQuAD 2.0 EM score.
Both methods perform poorly:
the calibrator (averaged over five choices of $\pknownoutlier$) achieves $54.0$ EM, while MaxProb achieves $53.1$ EM.\footnote{We evaluate on 4000 questions randomly sampled from the SQuAD 2.0 development set.} 
These results only weakly outperform the majority baseline of $48.9$ EM.

Taken together, these results indicate that identifying unanswerable questions is a very different task from knowing when to abstain under distribution shift.
Our setting focuses on test data that is dissimilar to the training data, 
but on which the original QA model can still correctly answer a non-trivial fraction of examples.
In contrast, unanswerable questions in SQuAD 2.0 look very similar to answerable questions, 
but a model trained on SQuAD 1.1 gets all of them wrong. 

\subsection{Changing ratio of in-domain to OOD}
\label{subsec:change_ratio}
Until now, we used $\alpha=\frac12$ both for $\dtest$ and training the calibrator.
Now we vary $\alpha$ for both, ranging from using only SQuAD to only OOD data (sampled from $\pknownoutlier$ for $\dcalibrate$ and from $\punknownoutlier$ for $\dtest$).

\begin{figure}[t]
\centering
\includegraphics[width=0.8\columnwidth]{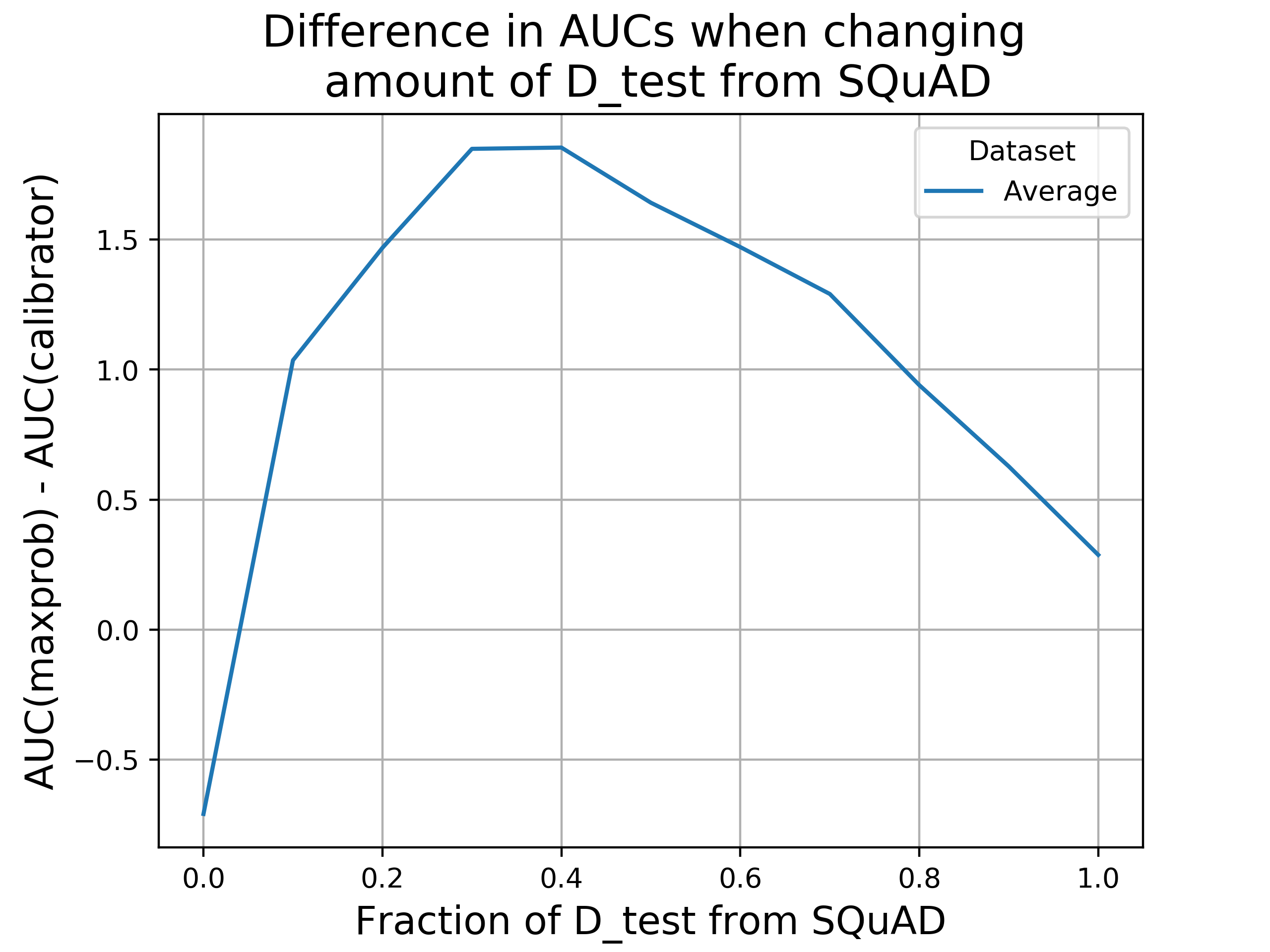}
\caption{
Difference in AUC between calibrator and MaxProb, as a function of how much of $\dtest$ comes from $\psource$ (i.e., SQuAD) instead of $\punknownoutlier$, averaged over 5 OOD datasets.
The calibrator outperforms MaxProb most when $\dtest$ is a mixture of $\psource$ and $\punknownoutlier$.
}
\label{fig:change_ratio}
\end{figure}

Figure \ref{fig:change_ratio} shows the difference in AUC between the trained calibrator and MaxProb.
At both ends of the graph, the difference is close to 0, showing that MaxProb performs well in homogeneous settings.
However, when the two data sources are mixed, the calibrator outperforms MaxProb significantly.
This further supports our claim that MaxProb performs poorly in mixed settings.  

\subsection{Allowing access to $\punknownoutlier$}
\label{sec:oracle}
We note that our findings do not hold in the alternate setting where we have access to samples from $\punknownoutlier$ (instead of $\pknownoutlier$).
Training the QA model with this OOD data and using MaxProb achieves average AUC of $16.35$, whereas training a calibrator achieves $17.87$;
unsurprisingly, training on examples similar to the test data is helpful.
We do not focus on this setting, as our goal is to build selective QA models for unknown distributions.

\section{Discussion}
In this paper, we propose the setting of selective question answering under domain shift,
in which systems must know when to abstain on a mixture of in-domain and unknown OOD examples.
Our setting combines two important goals for real-world systems: knowing when to abstain, and handling distribution shift at test time.
We show that models are overconfident on OOD examples, leading to poor performance in the our setting,
but training a calibrator using other OOD data can help correct for this problem.
While we focus on question answering, our framework is general and extends to any prediction task for which graceful handling of out-of-domain inputs is necessary.

Across many tasks, NLP models struggle on out-of-domain inputs.
Models trained on standard natural language inference datasets \citep{bowman2015large} generalize poorly to other distributions 
\citep{thorne2018fever,naik2018stress}.
Achieving high accuracy on out-of-domain data may not even be possible if the test data requires abilities that are not learnable from the training data \citep{geiger2019posing}.
Adversarially chosen ungrammatical text can also cause catastrophic errors \citep{wallace2019universal,cheng2020seq}.
In all these cases, a more intelligent model would recognize that it should abstain on these inputs.

Traditional NLU systems typically have a natural ability to abstain.
SHRDLU recognizes statements that it cannot parse, or that it finds ambiguous \citep{winograd1972language}.
QUALM answers reading comprehension questions by constructing reasoning chains, and abstains if it cannot find one that supports an answer \citep{lehnert1977process}.

NLP systems deployed in real-world settings inevitably encounter a mixture of familiar and unfamiliar inputs.
Our work provides a framework to study how models can more judiciously abstain in these challenging environments.

\paragraph{Reproducibility.} All code, data and experiments are available 
on the Codalab platform at \url{https://bit.ly/35inCah}.

\paragraph{Acknowledgments.} This work was supported by the DARPA ASED program under FA8650-18-2-7882.
We thank Ananya Kumar, John Hewitt, Dan Iter, and the anonymous reviewers for their helpful comments and insights.

\bibliography{refdb/all}
\bibliographystyle{acl_natbib}

\appendix
\section{Appendix}
\label{sec:appendix}
\subsection{Dataset Sources}
\label{sec:dataset_sources}
The OOD data used in calibrator training and validation was sampled from MRQA training data, 
and the SQuAD data for the same was sampled from MRQA validation data,
to prevent train/test mismatch for the QA model \citep{fisch2019mrqa}. 
The test data was sampled from a disjoint subset of the MRQA validation data.

\subsection{Calibrator Features and Model}
\label{sec:calibrator_features_model}
We ran experiments including question length and word overlap between
the passage and question as calibrator features. 
However, these features did not improve the validation performance of the calibrator. 
We hypothesize that they may provide misleading information about a given example,
e.g., a long question in SQuAD may provide more opportunities for alignment with the paragraph,
making it more likely to be answered correctly,
but a long question in HotpotQA may contain a conjunction,
which is difficult for the SQuAD-trained model to extrapolate to.

For the calibrator model, we experimented using an MLP and logistic regression. Both were slightly worse than Random Forest. 

\subsection{Outlier Detection for Selective Prediction}
\label{sec:outlier_detection}
In this section, we study whether outlier detection can be used to 
perform selective prediction. We train an outlier detector to detect
whether or not a given input came from the in-domain dataset (i.e., SQuAD)
or is out-of-domain,
and use its probability of an example being in-domain for selective prediction. The outlier detection model, 
training data (a mixture of $\psource$ and $\pknownoutlier$), 
and features are the same as those of the calibrator. We find that
this method does poorly, achieving an AUC of $24.23$, Coverage
at $80\%$ Accuracy of $37.91\%$, and Coverage at $90\%$ Accuracy of 
$14.26\%$. This shows that, as discussed in \refsec{related_but_distinct} 
and \refsec{main_results}, this approach is unable to correctly identify 
the OOD examples that the QA model would get correct.

\subsection{Underconfidence of MaxProb on SQuAD}
\label{sec:underconfidence_squad}
\begin{figure}[t]
\centering
\includegraphics[width=\columnwidth]{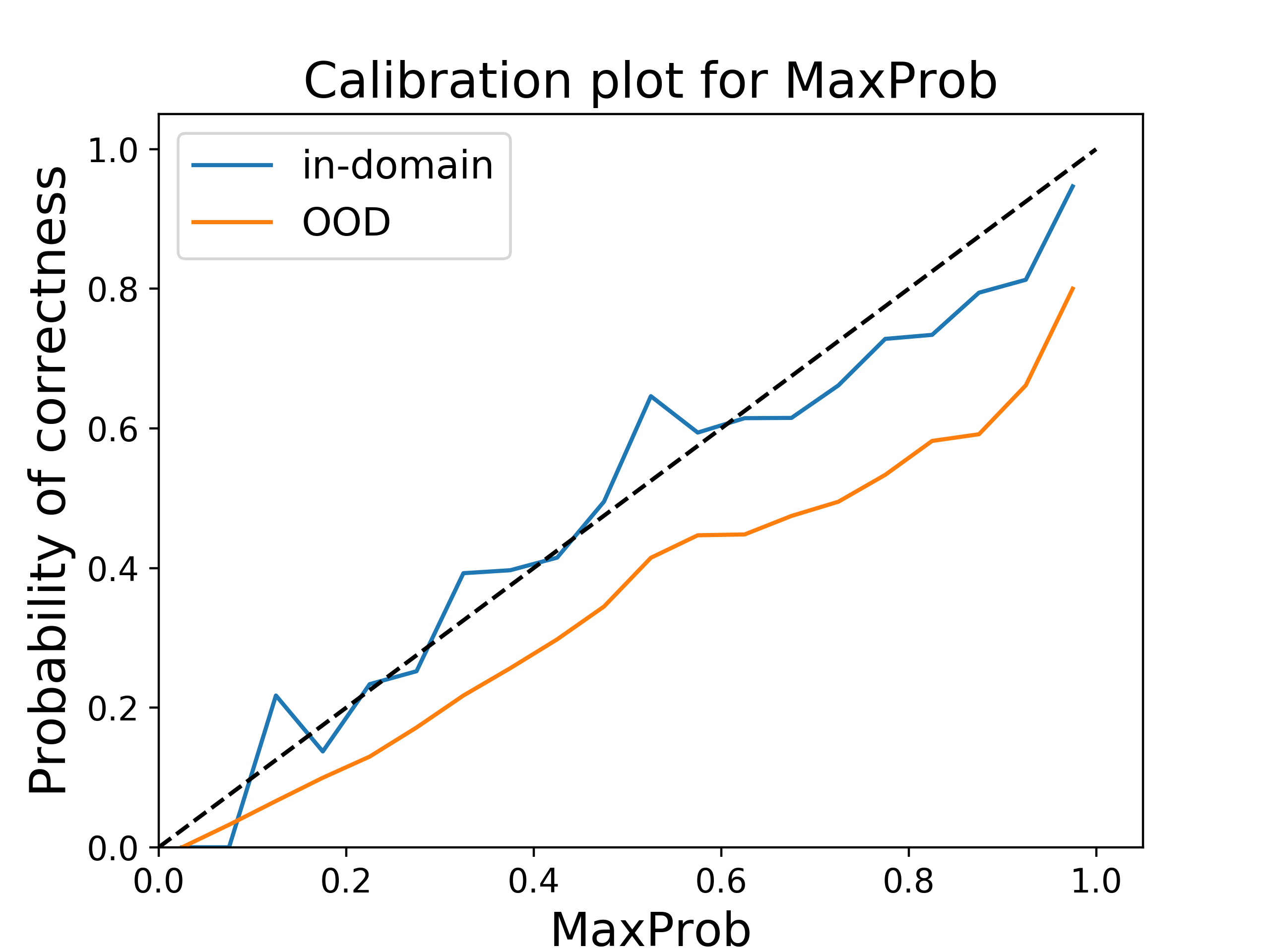}
\caption{
When considering only one answer option as correct,
MaxProb is well-calibrated in-domain, but is still overconfident out-of-domain.
}
\label{fig:strict_eval_maxprob}
\end{figure}

\begin{figure}[t]
\centering
\includegraphics[width=\columnwidth]{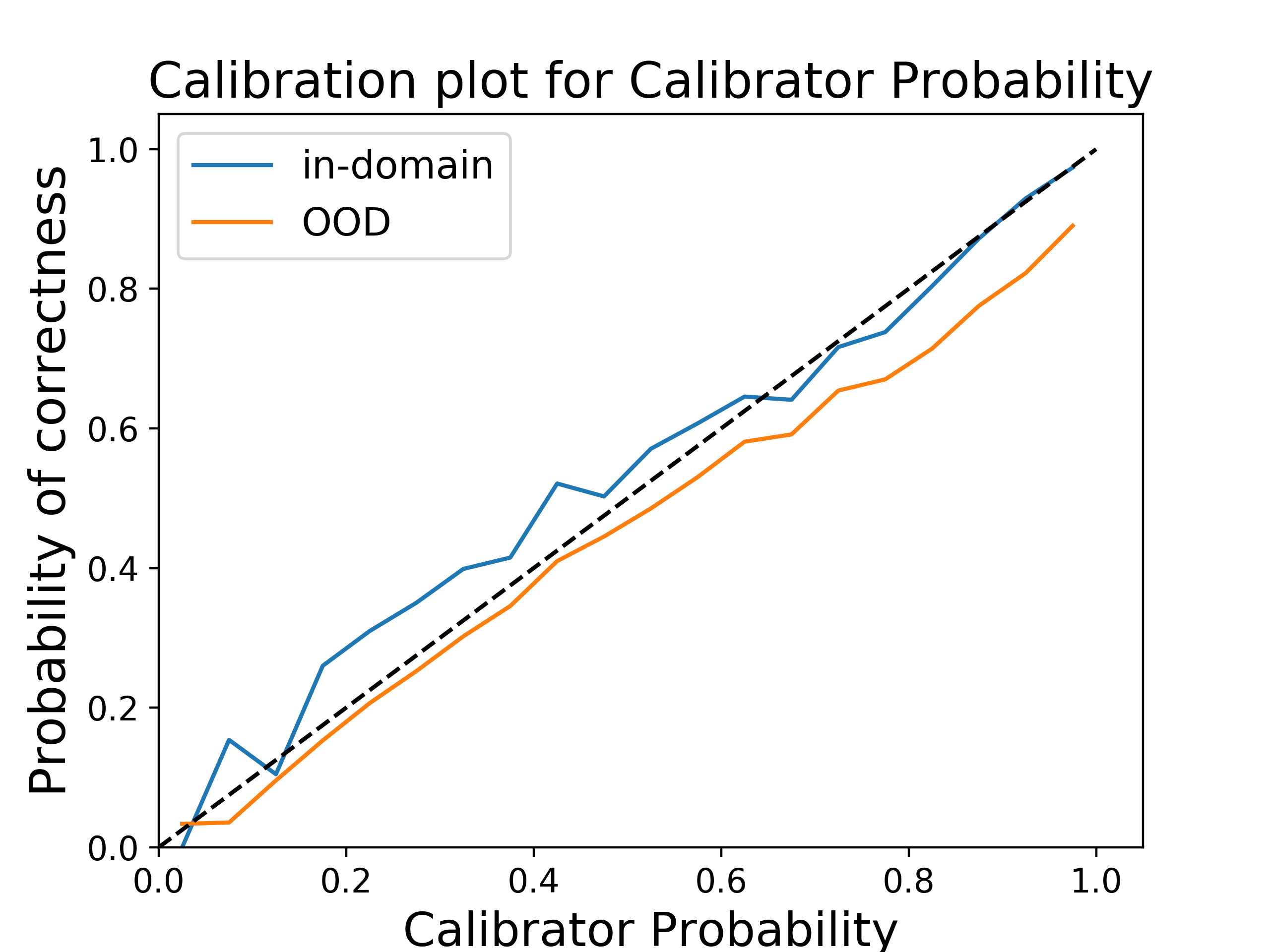}
\caption{
When considering only one answer option as correct,
the calibrator is almost perfectly calibrated on both in-domain and out-of-domain examples.
}
\label{fig:strict_eval_calibrator}
\end{figure}

As noted in \refsec{overconfidence}, MaxProb is 
underconfident on SQuAD examples due to the additional correct 
answer options given at test time but not at train time. 
When the test time evaluation is restricted to allow 
only one correct answer, we find that MaxProb 
is well-calibrated on SQuAD examples (Figure 
\ref{fig:strict_eval_maxprob}). The calibration of 
the calibrator improves as well (Figure 
\ref{fig:strict_eval_calibrator}).
However, we do not retain this restriction for the experiments, 
as it diverges from standard practice on SQuAD, and
EM over multiple spans is a better evaluation metric
since there are often multiple answer spans that are equally correct.

\subsection{Accuracy and Coverage per Domain}
\label{sec:per_domain_breakdown}
\reftab{main_table} in \refsec{main_results} shows the coverage of MaxProb and the calibrator over
the mixed dataset $\dtest$ while maintaining 80\% accuracy and 
90\% accuracy.
In \reftab{per_domain_table},
we report the fraction of these answered questions that are in-domain or OOD.
We also show the accuracy of the QA model on each portion.

Our analysis in \refsec{overconfidence} indicated that MaxProb was overconfident on OOD examples,
which we expect would make it answer too many OOD questions and too few in-domain questions. 
Indeed, at $80\%$ accuracy, $62\%$ of the examples MaxProb answers are in-domain, compared to $68\%$ for the calibrator.
This demonstrates that the calibrator improves over MaxProb by answering more in-domain questions,
which it can do because it is less overconfident on the OOD questions.

\begin{table}[h]
\resizebox{\columnwidth}{!}{%
\begin{tabular}{lcccc}
\toprule
\multicolumn{1}{c}{\textbf{}} & \textbf{\begin{tabular}[c]{@{}c@{}}MaxProb\\ Accuracy\end{tabular}} & \textbf{\begin{tabular}[c]{@{}c@{}}MaxProb\\ Coverage\end{tabular}} & \textbf{\begin{tabular}[c]{@{}c@{}}Calibrator\\ Accuracy\end{tabular}} & \textbf{\begin{tabular}[c]{@{}c@{}}Calibrator\\ Coverage\end{tabular}} \\ \midrule
\textbf{At 80\% Accuracy} & \multicolumn{1}{l}{} & \multicolumn{1}{l}{} & \multicolumn{1}{l}{} & \multicolumn{1}{l}{} \\
in-domain & 92.45 & 61.59 & 89.09 & \textbf{67.57} \\
OOD & 58.00 & 38.41 & 59.55 & \textbf{32.43} \\ \midrule
\textbf{At 90\% Accuracy} & & & & \\
in-domain & 97.42 & 67.85 & 94.35 & \textbf{78.72} \\
OOD & 71.20 & 32.15 & 72.30 & \textbf{21.28} \\
\bottomrule
\end{tabular}}
\caption{\label{tab:per_domain_table} Per-domain accuracy and coverage values of MaxProb and the calibrator ($\psource$ and $\pknownoutlier$) at 80\% and 90\% Accuracy on $\dtest$.} 
\end{table}

\end{document}